# Untargeted White-box Adversarial Attack with Heuristic Defence Methods in Real-time Deep Learning based Network Intrusion Detection System

Khushnaseeb Roshan[a,*], Aasim Zafar[a,b], Sheikh Burhan Ul Haque[a,c]

[a]*Department of Computer Science, Aligarh Muslim University, Aligarh, Uttar Pradesh, India*
[*]*Email: kroshan@myamu.ac.in*
[b]*Email: azafar.cs@amu.ac.in*
[c]*Email: sbuhaque@myamu.ac.in*

**Abstract**

Network Intrusion Detection System (NIDS) is a key component in securing the computer network from various cyber security threats and network attacks. However, consider an unfortunate situation where the NIDS is itself attacked and vulnerable; more specifically, we can say, "How to defend the defender?". In Adversarial Machine Learning (AML), the malicious actors aim to fool the Machine Learning (ML) and Deep Learning (DL) models to produce incorrect predictions with intentionally crafted adversarial examples. These adversarial perturbed examples have become the biggest vulnerability of ML and DL based systems and are major obstacles to their adoption in real-time and mission-critical applications such as NIDS. AML is an emerging research domain, and it has become a necessity for the in-depth study of adversarial attacks and their defence strategies to safeguard the computer network from various cyber security threads. In this research work, we aim to cover important aspects related to NIDS, adversarial attacks and its defence mechanism to increase the robustness of the ML and DL based NIDS. We implemented four powerful adversarial attack techniques, namely, Fast Gradient Sign Method (FGSM), Jacobian Saliency Map Attack (JSMA), Projected Gradient Descent (PGD) and Carlini & Wagner (C&W) in NIDS. We analyzed its performance in terms of various performance metrics in detail. Furthermore, the three heuristics defence strategies, i.e., Adversarial Training (AT), Gaussian Data Augmentation (GDA) and High Confidence (HC), are implemented to improve the NIDS robustness under adversarial attack situations. The complete workflow is demonstrated in real-time network with data packet flow. This research work provides the overall background for the researchers interested in AML and its implementation from a computer network security point of view.

## 1. Introduction

Every day, it becomes clear that ML and DL algorithms are accomplishing outstanding results in the domain where it is difficult to provide a set of rules for their proper functioning. Due to the availability of affordable computing power such as cloud services and GPU/TPU support, ML and DL are gaining more attention and providing promising results for future automation in almost every commercial sector like transportation [1], image recognition [2], speech recognition [3], healthcare [4], cyber security [5] [6] and so on. With such widespread application across multiple fields, some of its vulnerabilities are exploited by malicious hackers. Examples of these phenomena are the "Generalization" and "Robustness" capabilities of the model. The focus of the ML and DL research community for more than a decay is to optimize the model performance in terms of various performance metrics such as accuracy, precision, recall and reducing the false positive and negative cases [7] [8]. However, the robustness of the model cannot be ignored in today's era, as most of the ML and DL algorithms are vulnerable to adversarial attacks [9] [10]. AML is categorized into four groups: evasion, extraction, poisoning, and inference. In general, adversarial examples are generated by adding perturbation (small noise) into to actual input either in the training or testing phase, causing the model to misclassify the output. The other two classifications are black-box and white-box adversarial attacks.

In the black-box attack, the malicious actor has no knowledge of the targeted system. However, in the white-box attack, the model parameters are accessible to the hackers.

Moreover, to the best of our knowledge, most of the research studies in AML are focused on computer vision tasks [11] [12] [13] by the research community, and less is explored from the network and cyber security point of views such as in ML and DL based NIDS. The main concern is that performing adversarial attacks in network traffic datasets fundamentally differs from image datasets. In the case of the image dataset, the adversarial image seems different to the model even though it appears the same to the human observer. The Lp norms are used as distance metrics that exemplify the visual distance between two images. However, this concept fails in the computer network traffic domain as it is generally not practical to monitor network traffic data at the bit level, making it semantically different.

NIDS is used to monitor computer network traffic to detect intrusion or network anomalies over network hosts and servers. Traditionally, these systems are based on historical data, rule sets and signature-code created by human experts. However, this procedure is time-consuming. As a result, many solutions are implemented based on ML and DL algorithms in NIDS [14] [15] [16] [17] [18] [19], providing good automation to the security task. Furthermore, the unsupervised version of ML and DL algorithms provided great success in detecting unknown cyber-attack for which no previous signature is stored in the database [20] [21]. Despite the tremendous success of ML and DL algorithms in NIDS, adversarial machine learning raised a big security concern for the network security administrator. Adversarial attacks have the potential to fool ML and DL based NIDS, thus making computer networks more vulnerable to security threats [22] [23]. This research study is also motivated by the concept of online and streaming ML and DL methods [24] [25], where the model is trained with high speed and the latest network streaming data and updated regularly to maintain its prediction performance in real-time [26] [27]. This approach differs from the offline model-building procedure, therefore opening the door for the malicious user to inject adversarial samples during the online training and testing phase. Hence, ML and DL based systems are more vulnerable to online and streaming learning environments.

This research work revolved around exploring the sensitivity of DL based NIDS against widely known and powerful adversarial evasion attacks, namely: Fast Gradient Sign Method [28], Jacobian Saliency Map Attack [29], Projected Gradient Descent [30] and Carlini & Wagner. Further, we also examined and applied three heuristics defence methods, namely Adversarial Training, Gaussian Data Augmentation and High Confidence, to safeguard the NIDS against adversarial evasion attacks, thus analyzing a wide range of performance metrics. The findings of the extensive experimentation may help organizations and network security administrators to accomplish their goals with DL vulnerability analysis and its defence strategies successfully. Based on our findings, a cyber adversary with little insight into a target model (NIDS) might use the most effective perturbation method to attack NIDS successfully. And defenders might use the most suitable approach to defend NIDS against the adversarial evasion attack.

## 1.1. Main contributions

The novel contribution of the research work is as follows:

- It provides a detailed investigation and evaluation of four recent powerful adversarial perturbation techniques, namely FGSM, JSMA, PGD and C&W, from the network security perspective (NIDS). The latest and well-known CICIDS-2017 intrusion detection dataset is used to create adversarial perturbed samples with FGSM, JSMA, PGD and C&W methods to fool the NIDS with a high confidence score.

- Additionally, we explore three recent adversarial heuristics defence strategies, Adversarial Training, Gaussian Data Augmentation and High Confidence, to safeguard the NIDS against adversarial attacks.

- This research work provides the detailed technical description and intuition behind the adversarial attack approach in ML and DL based NIDS as well as its defence strategies to the researchers interested in the field of adversarial machine learning in the cyber security domain.
- The proposed research work is demonstrated in a real-time computer network with data packets flow from source to destination. The white-box and black-box, both adversarial attack scenarios, are conceptually explained. However, the white-box adversarial attack is experimentally implemented in DL based NIDS.

*1.2. Paper organization*

The overall structure of the research article is as follows: Section 1 is the Introduction section. It represents a brief overview of the requirement to increase the robustness and resilience of DL model against adversarial attacks followed by the main contribution of the research work. Section 2 is the prerequisite, it provides a brief idea of adversarial machine learning from its historical perspective. It also describes the broad categorization of adversarial attack and its defence methods. Section 3 is the literature review, it incorporates the recent study in adversarial ML and network security perspective. Section 4 provides the description of the dataset. Section 5 is the methodology section that explains the real-time adversarial attack execution in network traffic flow. The detailed description of black-box and white-box adversarial attacks are also explained in this section followed by model building and evaluation procedure with the adversarial attack and defence techniques. Section 6 discusses the experimentation results and findings. In Section 7, the limitations of the proposed work and future scope are discussed. Finally, Section 8 concludes this research work.

**2. Prerequisites**

This section briefly describes the taxonomy of adversarial machine learning and its common categorizations. We followed the simplistic approach for the sake of the reader's understanding. However, we would recommend the books [16] [17] to interested researchers for more technical details and exploration.

ML includes a wide number of techniques, such as extracting patterns, theory and analysis based on these algorithms. ML is defined as "A computer program is said to learn from experience E with respect to some class of tasks T and performance measure P, if its performance at tasks in T, as measured by P, improves with experience E" [31].

Adversarial machine learning is a research domain that converges the fields of ML, Computer Science, and Cybersecurity. The National Institute of Standards and Technology (NIST) [32] [33] defines it as "the process of extracting information about the behavior and characteristics of an ML system and/or learning how to manipulate the inputs into an ML system in order to obtain a preferred outcome." This area is broadly classified into four attack types, as outlined by NIST: evasion, poisoning, extraction, and inference [32] [33]. Figure 1 provides an illustrative representation of the categorization of adversarial machine learning.

- **Evasion Attacks:** Evasion attacks are designed to deceive an ML model into generating incorrect predictions. In the context of NIDS, these attacks trick the model into misclassifying network anomalies as benign traffic.
- **Poisoning Attacks:** Poisoning attacks take place during the training phase of an ML model. They involve injecting corrupt or malicious values into the input data. This manipulation can lead to the ML model being trained to accept malicious data as legitimate.
- **Model Extraction Attacks:** In model extraction attacks, hackers seek to acquire information about the ML model, including its hyperparameters and learning rates. This information is then used to construct a duplicate model that mimics the target system. This recreated model can subsequently be employed for malicious purposes.

- **Inference Attacks:** Inference attacks focus on gaining access to information regarding the training dataset. This is accomplished by leveraging the confidence scores associated with the model's predictions [34].

To facilitate understanding, these attack categories can be further dissected along three key dimensions: timing, information, and purpose:

- **Timing Dimension:** Evasion attacks are conducted after the ML model has been trained, during the testing phase. The objective is to manipulate the input data to deceive the model into producing erroneous results. In contrast, poisoning attacks occur during the model's training phase and aim to corrupt the training data to generate a compromised ML model.

- **Information Dimension:** This dimension pertains to the level of knowledge the attacker possesses about the ML model. White-box attacks occur when hackers possess comprehensive information about -the model, while black-box attacks transpire when the attacker has minimal or no knowledge about the learning algorithm. However, knowledge can be indirectly acquired through captured responses and queries.

- **Purpose Dimension:** Adversarial attacks can serve different objectives. Targeted attacks seek to manipulate the ML model into generating specific labels or targets. In contrast, untargeted or reliability attacks aim to diminish the model's confidence by increasing prediction errors.

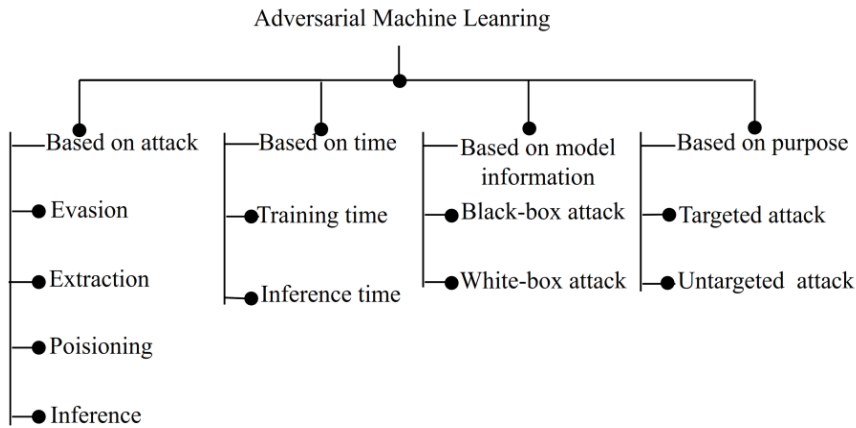

Figure 1. Categorization of adversarial machine learning.

## 2.1. Adversarial machine learning defence methods

Heuristic and certified defences are two broad categories of adversarial defence methods that protect ML and DL models from adversarial attacks [13]. Heuristic defences do not provide strong mathematical proof but are designed to detect or mitigate adversarial examples. The heuristic approach includes techniques like input preprocessing (e.g., input normalization), adversarial sample detection through outlier detection algorithms, adversarial training, etc. [13]. On the contrary, certified defence methods provide rigorous mathematical guarantees of a model's robustness against adversarial attacks. We highly recommend the research work proposed by Ren et al. [13] for further clarification on adversarial defence and attack methods. In our proposed research work, we have utilized three defence strategies. The first one is the Adversarial Training defence in which the model is retrained with the adversarial perturbation examples generated based on FGSM, JSMA, PGD and C&W. The second defence method is Gaussian Data Augmentation, in which random gaussian noise is added to the input data during inference time. The third is the High Confidence defence, which ensures that the model's predictions are only accepted when the model is highly confident in its output. This approach is relatively straightforward to implement and can provide an additional layer of protection against adversarial attacks. The detailed technical description of each method is available at [35].

The three utilized adversarial defence techniques are heuristic in nature and do not offer a strong guarantee against all types of adversarial attacks but are easier to implement and computationally efficient compared to certified defence methods. The effectiveness of these methods can be validated only through experiments. On the contrary, the certified or provable defence always maintains a certain accuracy under a well-defined class of attacks [36]. Some certified defence methods are Interval Bound Propagation [37], Certified Federated Adversarial Training [38] and Principled Adversarial Training [39]. However, in this research study, we have focused only on heuristics defence methods. Our future research study may explore certified adversarial defence methods against adversarial attacks from the NIDS perspective. Figure 2 shows the adversarial defence categorization as discussed above.

It is essential to understand that both approaches have their strengths and weaknesses, and their effectiveness can vary depending on the nature of the threats and the operational environment. In our case, the rationale behind exploring heuristics adversarial defence strategies lies in its adaptability to unknown threats, resilience against evolving evasion strategies, detection of threats without known signatures, flexibility, less complexity, and reduced false positives scenarios. On the contrary, certified defences can be vulnerable to these evasion strategies since they rely on known attack signatures and are less adaptive in nature. The certified defence methods excel at detecting known attacks with well-defined patterns but fail to detect unknown threats. However, it is essential to recognize that there is no one-size-fits-all solution, and a combination of both heuristic-based and certified defences may provide the most robust network security posture against a wide range of threats, including adversarial attacks in real-world scenarios [13]. We will explore hybrid defence strategies against adversarial attacks in our future research studies.

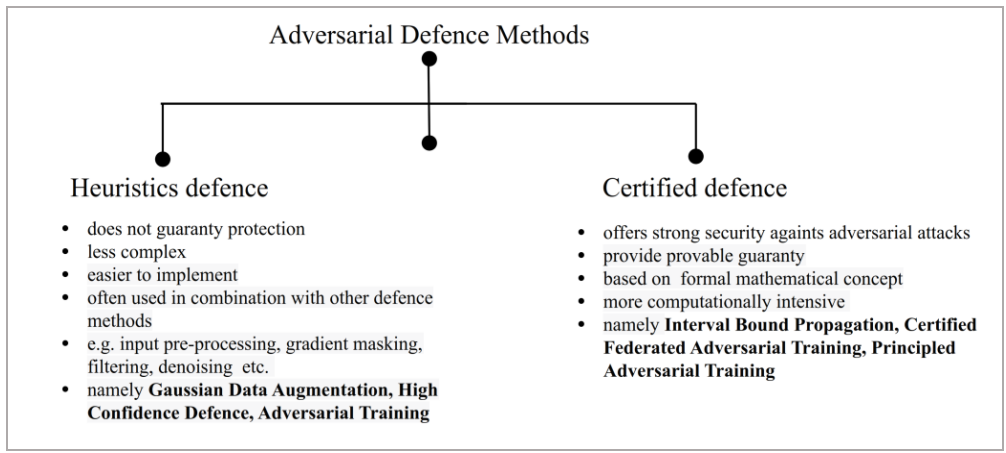

Figure 2. Categorization of adversarial defence methods

## 3. Literature Review

*3.1. Historical background*

The AML concept appeared initially in the 2000s. Dalvi et al. [40] presented a study in 2004 where they intentionally changed the email body to fool the spam classifier. In a later study, Lowd and Meek [41] applied adversarial classifier reverse engineering (ACRE) for spam email detection. The ACRE method is based on either continuous or boolean input data points and validates its use with actual spam filtering data. In 2006, Barreno et al. [42] presented several contributions, such as a taxonomy of various types of attacks on ML techniques, a range of defences against those attacks, and a discussion of concepts crucial to ML security. The author studied the broad question, "Can machine learning be secure?". Later, in 2010 [43], they further explored and clearly defined attacks and defence strategies. In 2013, Szegedy et al. [44] studied the intriguing properties of the DL algorithm (neural networks) and discovered that just by the addition of a small noise (perturbation) in the image, the DL model produces incorrect predictions with high confidence. They also explained that the characteristics of these perturbations are not just a random phenomenon, i.e., the same perturbations can lead to misclassification of the same input by a separate DL model with another subset of the same dataset.

Several research articles on AML have been published in the last ten years. For the interested reader, we would recommend the following survey articles. Liu et al. [45] conducted a comprehensive survey and explored the security threats from two perspectives: training and testing. In the training phase, they categorize defensive techniques into security assessment and countermeasure. In the testing phase, the two aspects are data privacy and data security. Akhtar and Mian [46] survey AML in computer vision tasks. The authors reviewed the contributions of the practical and real-world scenarios of adversarial attacks. Serban et al. [47] presented a more comprehensive survey that thoroughly covered the transferability concept and a wide range of adversarial attacks and their defence methods, and other research works are also recommended [48] [49].

*3.2. Unrestricted and restricted domain study*

Most of the research studies in AML are focused on unrestricted domains such as image and object classification [12] [13]. The main assumption of the adversary in the image domain is that the exploitation of each feature or pixel of the image is possible. However, the situation is different in the case of the constrained domain (e.g. network traffic dataset). This is because the feature value of the dataset can be binary, categorical or continuous. The features might also be correlated with each other, and some features may have fixed values and can not be modified in the adversarial examples generation phase. All the factors mentioned above raise another question: "Are constraint domains less susceptible to adversaries' generation ?". Sheatsley [50] empirically tested the mentioned hypothesis and experimented with the same over the constraints dataset. The author showed that the misclassification rate is more than 95% with adaptive JSMA and Histogram Sketch Generation (HSG) algorithm when applied to intrusion detection dataset. Some of the most effective and powerful adversarial algorithms are FGSM [28], JSMA [29], PGD [30] and C&W [51]. More technical details of the implemented algorithms are presented in the methodology section. The remaining portion of this section is recent related work applied in the NIDS with an adversarial machine learning perspective.

Pawlicki et al. [52] explored and analyzed the effect of FGSM, C&W, PGD and Basic Iterative Method (BIM) over various ML classifiers, namely, Artificial Neural Network (ANN), Random Forest (RF), AdaBoost and Support Vector Machine (SVM) in the context of network security. And later, they proposed a method based on neuron activation to mitigate its effects over the same ML classifiers. The subset of the CICIDS-2017 dataset is used for experimentation purposes. However, there is a scope for further enhancement in reducing the false positive scenario.

Wang [53] evaluated state-of-the-art adversarial algorithms in DL based NIDS in the context of the cyber security domain. The author analyzed and revealed that adversarial attacks, such as FGSM and JSMA, are used to evade DL based models in image classification tasks performed at various points of efficacy in the NIDS domain as well. The author evaluated the model performance with various evaluation metrics, i.e., precision, recall, accuracy, f1-Score, ROC curve, AUC score at the level of feature modification. The features that are altered specifically in JSMA algorithm are highly attractive to adversarial attacks. To be more specific, features that are most widely utilized imply a significant contribution to the vulnerability analysis of NIDS. These features require more attention from the security administrator to provide better protection and defensive efforts.

Guo et al. [54] analyzed the effect of BIM, a black-box adversarial attack, on the five ML based classifiers. The authors evaluated the effect of adversaries on Convolutional Neural Networks (CNN), SVM, k-Nearest Neighbor (KNN), Multilayer Perceptron (MLP), and the Residual Network (Resnet) with KDDCUP-99 and CICIDS-2017 two benchmark intrusion detection datasets. However, the author does not suggest any prevention mechanism to safeguard the model from adversaries.

Qureshi et al. [55] used the JSMA technique to attack the proposed RNN-ADV model. JSMA algorithm is more effective in identifying the features which can cause maximum change to the benign samples with minimum added perturbation. The author evaluated the proposed model with the NSL-KDD dataset with various performance metrics. However, the proposed work can be further extended with other complex adversarial and ML algorithms as well.

Alhajjar et al. [56] utilized Particle Swarm Optimisation (PSO), Genetic Algorithm (evolutionary computation), and Generative Adversarial Network (GAN) to generate adversarial examples to fool the ML and DL model in the domain of network security. Extensive experimentation is carried out over the eleven ML classifiers and with two publicly available datasets, i.e., UNSW-NB15 and NSL-KDD, to evaluate the performance of the proposed algorithm.

Usama et al. [57] used GAN to generate adversarial examples to evade the ML based NIDS. The study revealed that GAN could be used to thwart adversarial perturbation and make the model more robust against adversarial attacks. The functional behaviour of the network traffic is preserved by changing only non-functional attributes in the attack and defence phase. The proposed approach is evaluated over the KDDCUP-99 benchmark dataset and various ML algorithms, namely DNN, SVM, Decision Tree (DT), Random Forest (RF) etc. The performance of the GAN-based IDS is evaluated in terms of accuracy, precision, recall and f-score with four different phases, i.e., before the attack, after the attack, after adversarial training and after GAN-based adversarial training.

Clements et al. [58] studied the robustness of the DL-based NIDS Kitsune [59]. Kitsune is an ensemble and lightweight model based on the autoencoder for online network anomaly detection composed of various components, namely, Packet Capture, Feature-Extractor, Feature Mapper, and Anomaly Detector. The authors evaluated Kitsune on integrity and availability violations with four adversarial algorithms, FGSM, JSMA, C&W, and ENM, with Lp distance metrics and 1000 benign samples. Out of four algorithms, FGSM and JSMA performed with a significantly low success rate compared to C&W and ENM, specifically in availability attacks. The adversaries, in this case, know the target model and can directly generate the perturbed inputs for the DL model.

## 4. CICIDS-2017 Dataset

The latest CICIDS-2017 dataset [60], generated by the Canadian Institute for Cybersecurity, is used for experimentation. This dataset is publicly available in both PCAPs and CSV formats. It contains a variety of the most up to date inside and outside attacks, such as Port Scan, DDoS, DoS, Infiltration, Heartbleed, Botnet, Brute Force etc. The other benchmark datasets, such as KDDCUP-99 and its modified version NSL-KDD are very old and do not represent the real-network traffic environment. Hence CICIDS-2017 would be the most suitable and reliable dataset to evaluate the proposed approach efficiently. This dataset is labelled and contains a total of 79 features. Some features are based on statistical metrics like maximum, minimum, standard deviation etc., and others are packet flow, packet size distribution etc.

*4.1. Dataset preprocessing*

The subset of the CICIDS-2017 dataset is first preprocessed for any null and infinity values and filled with mean/median values. The dataset is normalized with the scikit-learn standard scaler function to efficiently train and test the model. The dataset is split into training, validation, and testing sets to train, validate, and test the DL based NIDS. The detailed description of the dataset is illustrated in Table 1. The dataset's training, validation and testing count is 102216, 34072 and 34072, respectively. Furthermore, the training, validation, and testing sample counts for binary classification (benign and attack) are (48557, 53659), (16014, 18058) and (43193, 47667), respectively. Figure 3 shows the dataset preprocessing procedure demographically.

Table 1 Dataset description

| Details | Counts |
| --- | --- |
| Training dataset samples | 102216 |
| Validation dataset samples | 34072 |
| Testing dataset samples | 90860 |
| Training benign and attack samples | ('BENIGN', 'Attack') – (48557, 53659) |
| Validation benign and attack samples | ('BENIGN', 'Attack') – (16014, 18058) |
| Testing benign and attack samples | ('BENIGN', 'Attack') – (43193, 47667) |

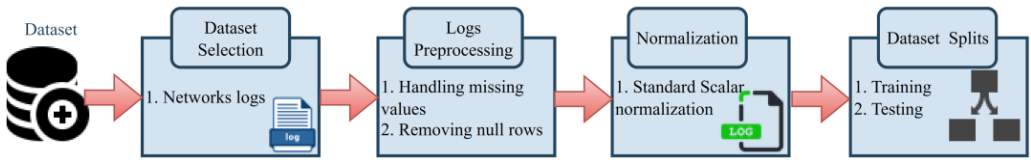

Figure 3. Dataset preprocessing steps

## 5. Methodology

This section describes the proposed research study in brief, real-time execution of the white-box and black-box adversarial attack on DL based NIDS in a computer network, the DL model architecture used to build NIDS, techniques used for adversarial attack generation and adversarial heuristics defence strategies. We have attempted to clarify the concept and each technical aspect of the proposed research work.

## 5.1. Proposed research study

In our study, we have conducted real-time adversarial attacks on Deep Learning-based Network Intrusion Detection Systems (NIDS), specifically employing untargeted white-box techniques such as Fast Gradient Sign Method (FGSM), Jacobian-based Saliency Map Attack (JSMA), Projected Gradient Descent (PGD), and Carlini and Wagner (C&W). Our empirical results demonstrate that the NIDS model lacks robustness in detecting adversarial perturbations, resulting in reduced model confidence scores across key performance metrics, including accuracy, precision, recall, F1-score, true positives, and false positives.

To enhance the NIDS's resilience against adversarial threats, we have implemented three heuristic defense strategies: Adversarial Training (AT), Gradient Domain Adversarial (GDA) training, and Heuristic Calibration (HC). Furthermore, we have conceptually discussed the applicability of black-box attacks in real-world implementations. While empirical validation of these black-box attacks is pending, it represents a significant direction for our ongoing and future research efforts. Our work aims to comprehensively address white-box adversarial attacks in the context of NIDS, as illustrated in Figure 4, which visually represents the key taxonomical terms and concepts employed in our study.

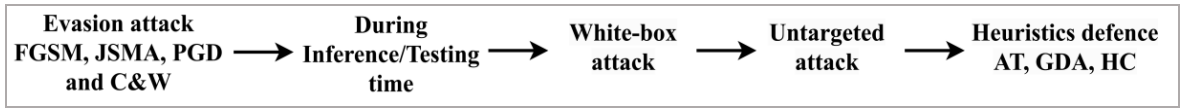

Figure 4. Taxonomical description used in the proposed research work

## 5.2. Real-time attack implementation

The real-time adversarial attack can be executed in two types. The first type is the white-box adversarial attack in which the hacker can access the NIDS model information. The hacker can use this information to create adversarial perturbation examples to fool the NIDS model into producing incorrect classifications. The second way is the black-box adversarial attack. In this case, the hacker does not have information of the target system (DL-NIDS) or dataset. To execute a black-box adversarial attack, the hacker has to use the concept of the surrogate model (that mimics the target model) to create adversarial perturbation examples that may deceive the actual NIDS model. This concept refers to the adversarial transferability phenomena discovered by Szegedy et al. [44]. It implies that adversarial examples crafted to deceive one model can also successfully deceive another, even if the model is trained on a different dataset [44]. Similarly, in another study, Papernot et al. [29] also revealed that adversarial examples generated by one model can effectively trick another DL model, even if the latter has a distinct architecture and different parameter configurations.

The execution of both white-box and black-box adversarial attacks in real-time is conceptually illustrated in the Figure 5. The complete procedure demonstrates the network packets flow from the source node to the destination node. The packets can choose different routing paths from source to destination. The hacker or adversary on the WAN can capture these packets to execute the adversarial attack on the target system (DL-NIDS). The complete approach is divided into three cases, namely: Case-1-No-Attack, Case-2-White-box and Case-3-Black-box Attack, as shown in Figure 5. We have explained each case in detail for further clarification. However, our main proposed research work that we empirically validated is Case-2 White-box adversarial attack. We have made the following assumptions to clarify the concept of real-time adversarial attack implementation.

**Assumptions**:
- Let's suppose we have four network packets: P1, P2, P3 and P4, moving from the source node to the destination node in the real-time computer network traffic.
- The hacker or the adversary can be present anywhere in the network. The adversary can access any router and packet information from WAN.
- Suppose packets P1 and P2 route through Path-3. Packet P3 routes through Path-1, and packet P4 routes through Path-2.
- Adversary-A has access to the targeted system, DL-NIDS, such as its parameters, architecture, and gradient details. Adversary-B does not have access to the DL-NIDS. This scenario represents the black-box adversarial transferability attack.

**Case-1- No Attack**

This case represents the clean path scenario in which no adversarial attack occurs through the adversary. The packet P4 moves from source to destination through Path-2 with routers R1-R5-R6 to the default gateway. The packet arrives at the DL-NIDS (installed on the default gateway) without the intervention of an adversary. The DL-NIDS pass this packet to the destination node.

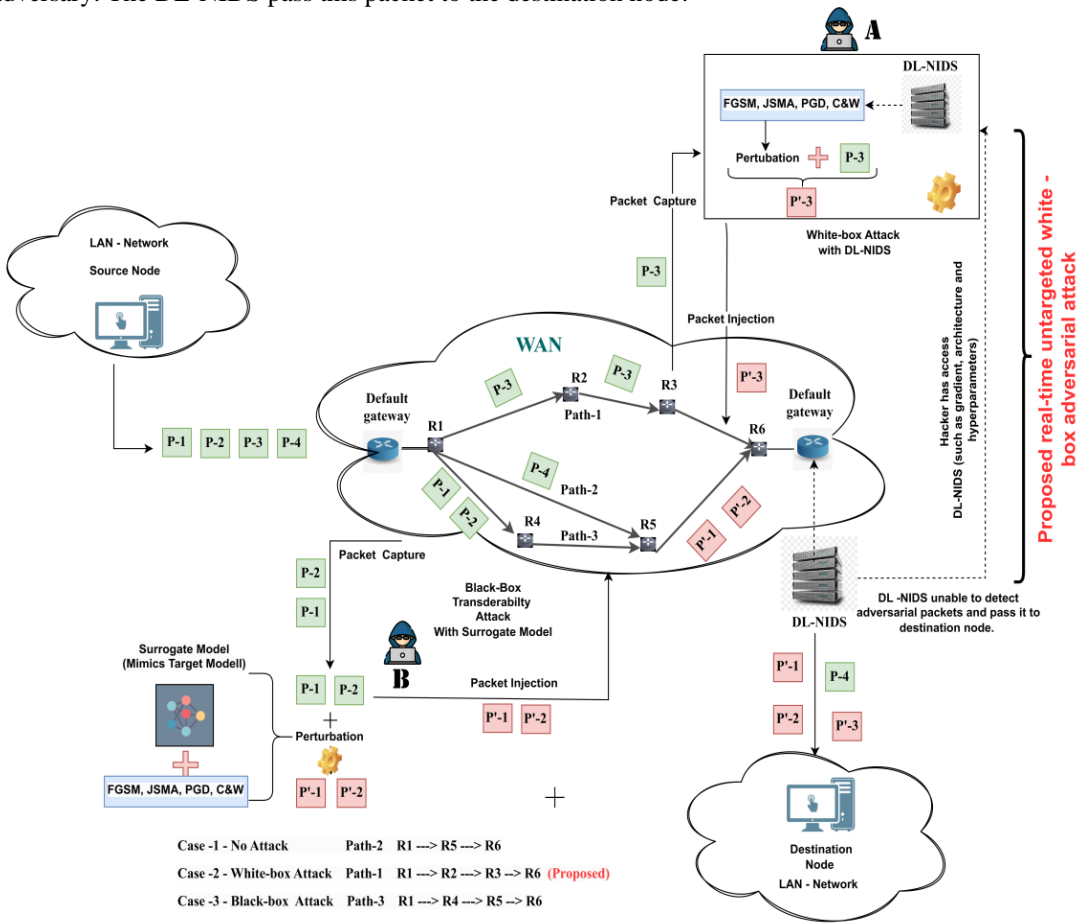

Figure 5. Real-time adversarial black-box and white-box attack execution in computer network

**Case-2- White-box Attack**

This case represents the white-box adversarial attack execution with FGSM, JSMA, PGD and C&W techniques on the packet P3, moving through Path-1 and routers R1-R2-R3-R6 to the default gateway. The targeted DL-NIDS is installed on the default gateway over the WAN. Adversary-A already has access to the DL-NIDS architecture information, such as gradients, hyperparameters, etc. Adversary-A utilizes this information to generate the adversarial perturbation packet. To execute the white-box attack, the adversary first captures the packet P3 from router R3. Adversary-A can use any of the adversarial perturbation generation techniques, namely FGSM, JSMA, PGD and C&W, to generate malicious perturbated packet P`3. The generated malicious packet is injected back into the same path in the network data stream. This malicious packet P`3 should be detected by the DL-NIDS at the default gateway. Due to the lack of robustness and resilience in detecting adversarial attacks, the DL-NIDS may allow malicious packets to pass through and reach their intended destination node.

**Adversarial packet generation algorithmic procedure**

Figure 6 illustrates the technical procedure for converting a benign network packet, denoted as P3, into an adversarial perturbed packet, represented as P'3. Each step in this process has been thoroughly described in Algorithm 1.

---

**Algorithm 1:** Adversarial Packet Generation Algorithm (using FGSM)

---

**Input:**
- P3: Captured benign network packet
- DL_NIDS: Deep Learning-based Network Intrusion Detection System (Classifier)
- ε (epsilon): Small noise factor for adversarial perturbation

**Output:**
- P'3: Adversarial perturbed packet

1. Extract features from the raw packet file P3 using CICIFlowMeter, organizing them into a feature matrix.
2. Reduce feature dimensionality using a feature reduction algorithm to optimize classifier performance.
3. Normalize feature values using min-max scaling.
4. Generate adversarial perturbation using the Fast Gradient Sign Method (FGSM):
   a. Define the FastGradientMethod() function with parameters DL_NIDS (Classifier) and ε (epsilon). This function is obtained from the adversarial-robustness-toolbox.
   b. Apply FastGradientMethod() to the normalized feature matrix to obtain the adversarial perturbation.
5. Create the adversarial packet P'3 by applying the adversarial perturbation to the original packet P3.
6. Forward the generated adversarial packet P'3 back into the same path in the network traffic flow.

---

The Adversarial Packet Generation Algorithm begins by extracting relevant features from the benign network packet (Step 1), organizing them into a feature matrix, and then optimizing classifier performance by reducing feature dimensionality (Step 2). After normalizing the feature values (Step 3), the algorithm employs the Fast Gradient Sign Method (FGSM) to generate adversarial perturbations (Step 4). FGSM calculates feature gradients and adjusts them based on a small noise factor (ε), resulting in a perturbed packet. The adversarial packet (P'3) is created by applying this perturbation to the original packet (P3). Finally, the adversarial packet is reintroduced into the network traffic flow, following its original path, in an attempt to evade intrusion detection by the DL-NIDS. This algorithm exemplifies the intricacies of adversarial attacks on network security systems, highlighting the need for robust intrusion detection mechanisms to counteract evolving threats effectively.

Furthermore, we would also want to highlight that the choice of the noise factor ε (epsilon) in adversarial attack generation is a critical parameter that directly influences the effectiveness and impact of these attacks on the deep learning based NIDS. To ensure the relevance and practicality of our adversarial attacks, we conducted an extensive empirical exploration of various ε values. Our primary goal was to find a balance that would allow us to expose vulnerabilities in the NIDS model while maintaining the imperceptibility of these perturbations in real-world network traffic. After thorough experimentation and analysis, ε = 0.003 emerged as the optimal choice. This value effectively fulfilled our research objectives by significantly impacting the NIDS model's performance, thereby highlighting its susceptibility to adversarial threats. Simultaneously, it ensured that the perturbations remained inconspicuous in typical network traffic scenarios.

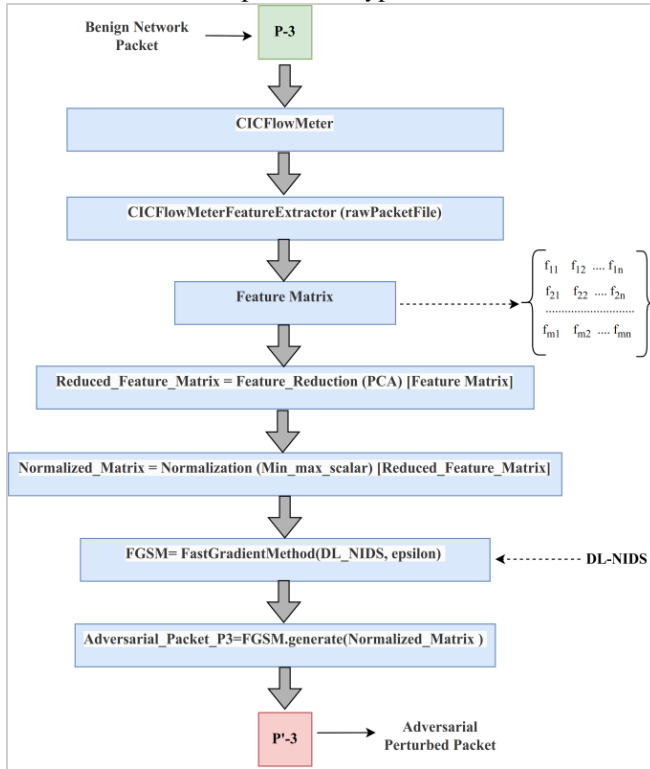

Figure 6. Adversarial packet generation algorithm process

**Case-3- Black-box Attack**

The black-box adversarial attack, also known as the transferability attack, is demonstrated through packets P1, P2. Both the packets route through Path-3 and routers R1-R4-R5-R6 to the default gateway. In this scenario, Adversary-B does not have access to the targeted DL-NIDS. Therefore, to execute the black-box transferability attack, the concept of the surrogate model that mimics the targeted DL-NIDS is utilized. The only difference here compared to the white-box attack is that the adversary has to use the surrogate model to generate an adversarial perturbated packet. In this research study, we have conceptually demonstrated this idea. We are already working on the empirical evaluation of the adversarial transferability concept from the cybersecurity perspective.

*5.3. DL- NIDS model*

A Deep Neural Network (DNN) is a computational model based on the Artificial Neural Network (ANN) comprising an input layer, one or more hidden layers and an output layer. Each layer consists of a number of artificial neurons fully connected with each other, making information flow from one layer to the next in a forward direction. DNN models are proven remarkable in learning hidden and abstract representations with the non-linear activation function (e.g. sigmoid, relu etc.) in complex and high-dimensional datasets such as computer network traffic.

In this research work, we used a supervised DL approach where the NIDS model is trained with the labelled dataset. Relu and Sigmoid, two activation functions, are used in hidden and output layers, respectively. The optimal hyperparameters of the model, such as learning rate, number of neurons, and number of hidden layers, are selected based on the random search algorithm [61], which converges fast compared to grid search algorithms [61], if the semi-optimal parameters are known. Algorithm 2 shows the evaluation of the originally built model without any adversarial attack over the clean dataset.

---

**Algorithm 2** : Model training and testing evaluation

1. **Input**: training_data, test_data, model
2. **Output**: CR, CM, ROC

   # list of variables
3. true_label ← test_data (true_label)

   # model evaluation
4. model ← model.fit(training_data)
5. predict ← model.predict(test_data)
6. CR ← classification_report (predict, true_label)
7. CM ← confusion_matrix (predict, true_label)
8. ROC ← roc_curve (predict, true_label)
9. **return** CR CM ROC

---

*5.4. Adversarial attack generation algorithms*

*5.4.1. Fast Gradient Sign Method (FGSM)*

Goodfellow et al. [28] proposed FGSM, an untargeted attack method and explained that even a simple ML model is vulnerable to adversarial attack if its input has sufficient dimensionality. FGSM is a rapid approach to generate adversarial samples with a small perturbation (noise) that can fool various ML models into producing incorrect predictions. FGSM algorithm optimizes the $L_P$ norm by taking one step in the opposite direction of the gradient to each element of input vector *x*. The optimal max-norm constrained perturbation is formulated as in Equation 1. The complete procedure is described in Algorithm 3.

$$\eta = \epsilon\, sign(\nabla_x J(\Theta, x, y)) \tag{1}$$

Where *x* is the input to the model, *y* is the target output vector, $\Theta$ represents the model's parameter, and $\nabla_x J(\Theta, x, y)$ is the gradient of the loss function of the model trained with $J(\Theta, x, y)$ cost. And $\epsilon$ is used to control the perturbation magnitude (usually very small).

*5.4.2. Jacobian Saliency Map Attack (JSMA)*

Papernot et al. [29] proposed JSMA which is based on $L_0$ norm. This method is based on the Jacobian Matrix, used to calculate the forward derivative of the loss function *f*, learned by the model during the training phase. JSMA approach iteratively calculates the saliency map to minimize the $L_0$ norm. In this way, the method identifies the most significant feature of the input vector *x* that triggers a large variation in the model predictions. The derivative over the input sample *x* is formulated as in Equation 2.

$$J_f(x) = \partial f(x)/\partial x = [\partial f_j(x)/ \partial x_i]_{i \times j} \quad (2)$$

*5.4.3. Project Gradient Descent (PGD):*

Madry et al. [30] proposed PGD, which is the extended version of the Basic Iterative Method (BIM) [62] and is based on the universal first-order $L\infty$ attack. PGD method iteratively searches for the perturbation and optimizing saddle point (min-max) formulation. The authors focus on two main questions based on prior literature work in adversarial machine learning. The first is "how to generate strong adversarial examples that fool the model with high confidence and require only small perturbations." The second is " How to train a model in such a way that no adversarial examples are possible or adversaries can not find them". The authors demonstrated that if the model is robust against the PGD adversaries, it will also defend against other adversarial attacks encompassing similar approaches ($L\infty$ norm). The problem formulation is as follows in Equation 3.

$$x^0_{adv} = x,$$

$$x_{(n+1)} = Clip_{x,\epsilon} \{x_n + \epsilon\ sign(\nabla_x J(\Theta, x, y))\} \quad (3)$$

*5.4.4. Carlini and Wagner (C&W):*

Carlini and Wagner proposed the C&W adversarial technique [51], a powerful targeted attack formulated based on the optimization problem and can restrict $L_0$, $L_2$ and $L_{inf}$ norm to make perturbation imperceptible. The authors demonstrated that the attack could defeat defensive distillation [63], a technique that reduces adversarial sample generation in DNN. The core objective function is shown in Equation 4.

$$\min_{\delta}\ ||\ \delta\ ||_p + c.f(x + \delta) \quad (4)$$

Here, $\delta$ represents the adversarial perturbation. It is the difference between the actual input and the adversarial sample. A modified binary search is used to find constant c, and *f(.)* is the objective function. The authors suggested seven possible choices of the objective function, and one of the functions is presented in Equation 5.

$$f(x') = max(max\ \{Z(x')i : I \neq t\} - Z(x')t, -k) \quad (5)$$

Here, *Z(.)* denotes the softmax loss function of the model, k is the constant that controls the confidence of the incorrect predictions.

**Algorithm 3** Model evaluation after adversarial attacks
1. **Input:** test_data, model
2. **Output:** CR, CM, ROC

    # list of variables
3. true_label ← test_data(true_label)
4. attack_algorithms ← {fgsm, jsma, pgd, c&w}
5. adv_list ← {}

    # generate adversarial examples with FGSM, JSMA, PGD and C&W
6. **for** algo **in** attack_algorithms **do**:
7.    adv_list.append ← generate_adversarial_samples(test_data, model, algo)
8. **end for**

    # evaluate model over generated adversarial samples for each method
9. **for** list **in** adv_list **do**:
10.    predict ← model.predict(list)
11.    CR ← classification_report (predict, true_label)
12.    CM ← confusion_matrix (predict, true_label)
13.    ROC ← roc_curve (predict, true_label)
14.    **return** CR CM ROC
15. **end for**

## 5.5. Adversarial attack defence algorithms

This section includes the defence strategies applied to prevent adversarial attacks in detail. It mainly includes Adversarial Training, Gaussian Data Augmentation and High Confidence defence. The complete procedure is described in Algorithm 4.

### 5.5.1. Adversarial Retraining (AT):

Adversarial training is a natural and intuitive defence against adversarial attacks. This method increases the robustness of the model by injecting the adversarial samples during the training procedure. It is one of the most effective defences against adversarial attacks and achieved state-of-the-art results on benchmark datasets. In the proposed work, we used the adversarial ensemble training method (EAT) [64], which increases the diversity of the adversarial samples in the training phase, thus enhancing the robustness of the model against transferred adversaries as well. It is experimentally demonstrated that the adversarial trained model is robust against both single and multi-step attacks generated on the models.

The authors followed the work presented in [30] and [44] and used a variant of Empirical Risk Minimization (ERM), which minimizes the risk of adversarial samples. The ERM function is formulated as:

$$h^* = argmin\ E_{(x,y) \sim D}\ [\ max\ _{||x^{adv} - x||_\infty \le \epsilon}\ L(h(x^{adv}), y)] \tag{6}$$

$x$ is the original input vector, $x^{adv}$ is its adversarial version, $y$ is the true label. The given formulated $h^*$ is the combination of inner maximization and outer minimization problems. The author uses both the clean $x$ and the adversarial samples $x^{adv}$ in the training procedure.

### 5.5.2. Gaussian Data Augmentation (GDA):

In general, the adversarial training method provides defence against white-box attacks and fails to defend against black-box attacks effectively [64]. Adversarial training method increases the model robustness in a few directions of the gradient points only. Hence the model can easily be fooled and provide incorrect predictions in other directions. In contrast, the Gaussian data augmentation defence method [65] allows exploration in multiple directions, smoothing the model's confidence. Moreover, the Gaussian method is peculiar to use Gaussian distribution in generating perturbation samples, which encourages the model to gradually reduce its confidence when moving away from the input sample. The author formulated the problem as shown in Equation 7.

$$\min_\theta \; E_{(x,y) \sim D} \; E_{\Delta x \sim N(0,\sigma^2)} \; J(\theta, x + \Delta x, y) \qquad (7)$$

$$\min_\theta \; E_{(x,y) \sim D} \; 1/N \; E_{(i=1 \text{ to } N)} \; [J(\theta, x + \Delta x_i, y)] \qquad (8)$$

Here, σ corresponds to the acceptable non-perceivable perturbation. The objective is to make sure that the posterior distribution p(y|x) follows N($x$, $\sigma^2$). And, later the authors use a Monte Carlo approximation of the previous problem, as shown in Equation 8.

---

**Algorithm 4** Model evaluation after adversarial defence

1. **Input:** test_data, adv_list, model
2. **Output:** CR, CM, ROC
   #list of variables
3. defence_models ← {}
   #generate models after applying three adversarial defence algorithms
4. ADV_model ← AdversarialTrainer (model, adv_list, other_param)
5. GS_model ← GaussianDataAugmentation (model, other_param, clip_values)
6. HC_model ← HighConfidence (model, other_param, clip_values)
7. defence_models ← {ADV_model, GS_model, HC_model}

   # evaluate updated models performance under the adversarial attacks
   # (i.e. FGSM, JSMA, PGD and C&W)
8. **for** model **in** defence_models **do**:
9.   **for** list **in** adv_list **do**:
10.     predict ← model.predict(list)
11.     CR ← classification_report (predict, true_label)
12.     CM ← confusion_matrix (predict, true_label)
13.     ROC ← roc_curve (predict, true_label)
14.     **return** CR CM ROC
15.   **end for**
16. **end for**

---

### 5.5.3. High Confidence (HT):

High Confidence defence [66] is based on the postprocessor defence approach, which selects high confidence prediction to return as the model prediction. The postprocessor method does not include the loss function evaluation, class gradient or the calculation of loss gradients. The method is demonstrated in the IBM adversarial robustness toolbox defence mechanism [66].

Figure 7 shows the demographic representation of the adversarial attack and defence methods used in the proposed research study. It shows the abstract procedure of detection of the adversarial examples input into the model during the inference phase. And the abstract procedure of defence method to increase the resilience of the model against adversarial attacks. The study demonstrates that our proposed NIDS model exhibits a pro-robust behavior when confronted with adversarial attacks. Pro-robustness in this context signifies that our model consistently maintains its high performance and accuracy levels, even in the presence of sophisticated adversarial attack techniques such FGSM, JSMA, PGD and C&W. The effectiveness of NIDS model's pro-robustness is evidenced by its ability to withstand adversarial perturbations while still providing reliable intrusion detection in network traffic data.

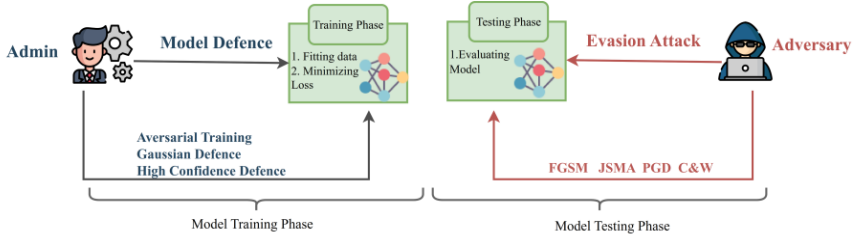

Figure 7. Block diagram of model attack and defence methods

## 6. Results and discussion

This experiment is carried out on Window-11 Operating System, Core i7 processor, 500 SSD, and 16 GB DDR4L RAM Support. The python 3.7 and other supportive libraries like Keras, Pandas, Scikitlearn, Matplotlib etc., and Jupyter Notebook is used as IDE for experimentation.

*6.1. Evaluation metrics*

The proposed NIDS model is evaluated on the following metrics.

**Confusion Metrics (CF):** It is divided into four groups True Positives(TP), True Negatives(TP), False Positives (FP) and False Negatives (FN). TP and TN represent the correct prediction of the attack and benign samples. FP and FN represent the incorrect identification of attack and benign samples. The other performance measures based on the CF are Accuracy (AC), Precision(P), Recall(R) and F-score (F), described as follows:

**Accuracy (ACC):** it is defined as the ratio between the correct samples to all the samples predicted by the classifier.

$$ACC = TP + TN / TP + TN + FP + FN \qquad (9)$$

**Precision (P):** It is defined as the ratio between the correct positive samples to all the positive samples predicted by the classifier.

$$Precision\ (P) = TP / TP + FP \qquad (10)$$

**Recall (R):** is the ratio between the true positive samples predicted by the classifier to all the actual positive samples. It is also known as True Positive Rate (TPR).

$$Recall\ (R) = TP\ /\ TP + FN \qquad (11)$$

**F-Score (F):** represents the harmonic mean of Precision and Recall.

$$F\text{-}Score = 2 * R * P\ /\ R + P \qquad (12)$$

**False Positive Rate (FPR):** it is defined as the ratio between the falsely predicted positive samples to the actual negative samples.

$$False\ Positive\ Rate\ (FPR) = FP\ /\ FP + TN \qquad (13)$$

*6.2. Performance Evaluation*

We initiated our performance evaluation of the NIDS model by examining clean data, which we designated as the "pre-attack evaluation" section. Subsequently, we exposed the NIDS to various adversarial attacks and meticulously assessed its performance under these conditions in the "post-attack evaluation" section. Finally, we extensively evaluated the model's performance after implementing our proposed adversarial defense approaches, which constituted the "post-defense evaluation" section.

*6.2.1. Pre-attack Evaluation*

The original dataset is split into the ratio of 60:40 percentage for NIDS model training and testing purpose. The validation split factor is set to 0.2 during the model training phase. The optimized architecture for the NIDS model on the CICIDS-2017 dataset consists of four hidden layers, with 60, 40, 20 and 10 neurons in each layer with an optimal batch size of 8192. The l2 regulariser is set to 0.0001, relu activation function is used for hidden layers, and the sigmoid function is for the output layer. The model is trained and validated over training and validation datasets over the 50 epochs. The NIDS model training & validation accuracy, training & validation loss are shown in Figure 8.

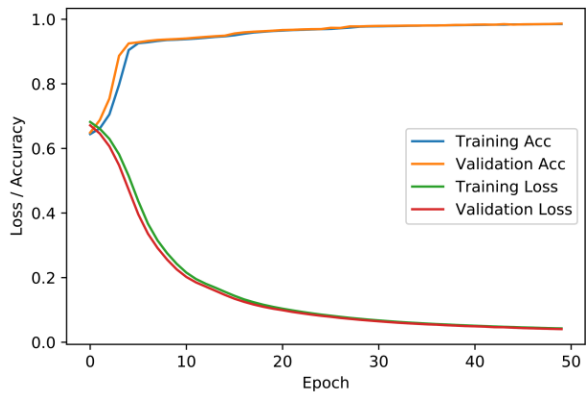

Figure 8. Model loss and accuracy curve

The obtained results of the NIDS model on the testing dataset are summarised in terms of classification report and confusion matrix, as shown in Table 2 and Table 3. The weighted accuracy precision, recall, and f-score obtained by the model are 98.54%, 98.56%, 98.54%, and 98.54%, respectively. The obtained AUC score is 98.55. The results on the clean dataset (without adversarial attack) are quite acceptable in the case of the supervised model. The following subsection analyzed the same NIDS model under the four adversarial attack algorithms.

Table 2 Classification report

| Label | AC | P | R | F |
|---|---|---|---|---|
| 0 |  | 99.71 | 97.21 | 98.44 |
| 1 | 98.54 | 97.52 | 99.74 | 98.62 |
| MA |  | 98.62 | 98.47 | 98.53 |
| WA |  | 98.56 | 98.54 | 98.54 |

Table 3 Confusion matrix

| LB |  | Predicted | |
|---|---|---|---|
|  |  | 0 | 1 |
| Actual | 0 | 41986 | 1207 |
|  | 1 | 122 | 47545 |

### 6.2.2. Post-attack evaluation

This section investigated and summarised the results of NIDS model under the adversarial attack situation on four algorithms, namely, FGSM, JSMA, PGD and C&W, in detail with various performance metrics.

#### 6.2.2.1. Results under FGSM

In the case of the FGSM algorithm, the perturbation noise is set to 0.003 with a batch size of 1000. The gradient sign method is used to produce adversarial perturbed samples. The min and max value of the output vector lies between -0.98 to +0.99. We followed the untargeted approach of the FGSM methods in this experiment. We can see a significant performance degradation of the NIDS model in terms of classification report, confusion matrix, and AUC score. The reduced weighted accuracy, precision, recall, f-score, and AUC score under the untargeted FGSM attack are 57.59%, 66.93%, 57.59%, 52.54% and 59.232, respectively, as shown in Table 4 and Figure 10. The misclassification rate has also increased. The predicted false positive and false negative scenarios are (3182, 35354), as listed in Table 5.

#### 6.2.2.2. Results under JSMA

JSMA algorithm is based on a Jacobian Matrix as described in [29]. To generate adversarial examples, the theta and gamma parameters are set to 0.03, 0.02 with a batch size of 1000. The JSMA algorithm attempts to search the feature space to find the most significant features that have a maximal impact on the output. And, during each iteration, the feature space size reduces, hence improving its performance. In the JSMA attack, the reduced weighted accuracy, precision, recall, f-score, and AUC score are 66.58%, 7.66%, 66.58%, 63.66% and 68.037, respectively, as shown in Table 4 and Figure 10. The predicted false positive and false negative scenarios are (1051, 29312), as listed in Table 5.

*6.2.2.3. Results under PGD*

PGD is a strong attack algorithm based on the saddle points concept and a definitive method for constrained large-scale optimization method [30]. PGD is a multi-step attack in contrast to FGM. In our case, to generate the perturbation examples, the noise parameter is set to 0.003. The other parameter, like step size, iteration and batch size, is set to 10, 100, and 1000, respectively. The reduced weighted accuracy, precision, recall, f-score and AUC score are 56.81%, 66%, 56.81%, 51.47% and 58.485, respectively, as shown in Table 4 and Figure 10. The predicted false positive and false negative scenarios are (3273, 35966), as listed in Table 5.

*6.2.2.4. Results under Carlini & Wagner*

The C&W attack is another powerful adversarial attack based on the L-norm (0, 2 and inf). In our case, the L-2 norm C&W algorithm is used to generate perturbed adversarial examples. The learning rate is set to 0.02. The parameters like binary search step and max iteration are both set to 10. The other variables, initial const, max_halving, and max_doubling, are fixed to 0.01, 5 and 5, respectively, with a batch size of 1000. As a result, the reduced weighted accuracy, precision, recall, f-score and AUC score are 61.74%, 74.82%, 61.74%, 57.11% and 63.41, respectively, as shown in Table 4 and Figure 10. The predicted false positive and false negative scenarios are (1182, 33578), as listed in Table 5. For ease of understanding and visibility, Figure 9 graphically depicts the classification report before and after the FGSM, JSMA, PGD and C&W attack on the NIDS model.
s

Table 4 Classification report after the adversarial attack

| LB | FGSM | | | | JSMA | | | |
|---|---|---|---|---|---|---|---|---|
| | AC(%) | P(%) | R(%) | F(%) | AC(%) | P(%) | R(%) | F(%) |
| 0 | 57.59 | 53.09 | 92.63 | 67.5 | 66.58 | 58.98 | 97.57 | 73.52 |
| 1 | | 79.46 | 25.83 | 38.99 | | 94.58 | 38.51 | 54.73 |
| MA | | 66.28 | 59.23 | 53.24 | | 76.78 | 68.04 | 64.12 |
| WA | | 66.93 | 57.59 | 52.54 | | 77.66 | 66.58 | 63.66 |
| LB | PGD | | | | C&W | | | |
| 0 | 56.81 | 52.61 | 92.42 | 67.05 | 61.74 | 55.58 | 97.26 | 70.74 |
| 1 | | 78.14 | 24.55 | 37.36 | | 92.26 | 29.56 | 44.77 |
| MA | | 6537 | 5848 | 52.2 | | 73.92 | 63.41 | 57.75 |
| WA | | 66 | 56.81 | 51.47 | | 74.82 | 61.74 | 57.11 |

Table 5 Confusion matrix after the adversarial attack

| FGSM | | | | JSMA | | | |
|---|---|---|---|---|---|---|---|
| LB | | Predicted | | LB | | Predicted | |
| | | 0 | 1 | | | 0 | 1 |
| Actual | 0 | 40011 | 3182 | Actual | 0 | 42142 | 1051 |
| | 1 | 35354 | 12313 | | 1 | 29312 | 18355 |
| PGD | | | | C&W | | | |
| LB | | Predicted | | LB | | Predicted | |
| | | 0 | 1 | | | 0 | 1 |
| Actual | 0 | 39920 | 3273 | Actual | 0 | 42011 | 1182 |
| | 1 | 35966 | 11701 | | 1 | 33578 | 14089 |

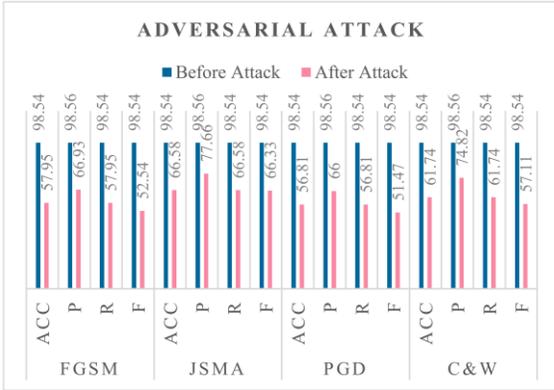 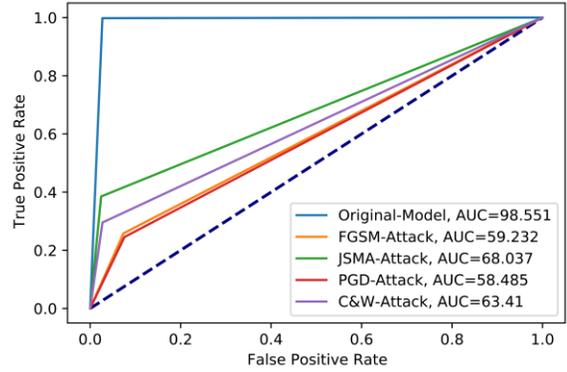

Figure 9. Original vs adversarial attack classification report     Figure 10. Original vs adversarial attack ROC curve

### 6.2.3. Post-defense evaluation

This section describes three defence mechanisms we opted against adversarial attacks. The detailed analysis and investigation are presented in the following subsection.

#### 6.2.3.1. Results after adversarial training defence

As discussed in the methodology section, in the adversarial defence technique, the model is retrained by augmenting training data along with adversarial perturbed examples to boost its robustness. Adversarial training defence is one of the most intuitive and effective approaches against adversarial attacks. In our case, the NIDS model has been retrained along with the four different adversarial perturbed examples generated with FGSM, JSMA, PGD and C&W techniques separately with 100 epochs. The retrained model is then evaluated using the previous adversarial perturbed examples. As a result, the NIDS model is significantly improved in terms of classification reports, confusion matrix, and AUC score, as shown in Table 6, Table 7, Figure 12. For better understanding and visibility Figure 11 demographically comparing the classification report of the original model in an ideal situation and under the adversarial attack situation in all the four cases, i.e., FGSM, JSMA, PGD and C&W.

The improved weighted accuracy, precision, recall and f-score and AUC score under the FGSM, JSMA, PGD and C&W adversarial attack are, (98.7%, 98.72%, 98.7%, 98.7%, 98.64), (98.47%, 98.51%, 98.47%, 98.47%,98.402), (98.68%, 98.71%, 98.68%, 98.68%, 98.627) and (71.56%, 80.35%, 71.56%,69.86%, 72.797), respectively. The improved classification report shows that false positive and false negative scenarios, under the FGSM, JSMA, PGD and C&W approach, have been reduced to (1083, 101), (1303, 85), (1094, 102) and (907, 24933), respectively, with the adversarial training defence.

Table 6 Classification report after adversarial defence

| | FGSM | | | | JSMA | | | |
|---|---|---|---|---|---|---|---|---|
| LB | AC(%) | P(%) | R(%) | F(%) | AC(%) | P(%) | R(%) | F(%) |
| 0 | | 99.76 | 97.49 | 98.61 | | 99.8 | 96.98 | 98.37 |
| 1 | 98.7 | 97.77 | 99.79 | 98.77 | 98.47 | 97.33 | 99.82 | 98.56 |
| MA | | 98.77 | 98.64 | 98.69 | | 98.57 | 98.4 | 98.47 |
| WA | | 98.72 | 98.7 | 98.7 | | 98.51 | 98.47 | 98.47 |
| LB | PGD | | | | C&W | | | |
| 0 | | 99.76 | 99.47 | 98.6 | | 62.91 | 97.9 | 76.6 |
| 1 | 98.68 | 97.75 | 99.79 | 98.76 | 71.56 | 96.16 | 47.69 | 63.76 |
| MA | | 98.76 | 98.63 | 98.68 | | 79.54 | 72.8 | 70.18 |
| WA | | 98.71 | 98.68 | 98.68 | | 80.35 | 71.56 | 69.86 |

Table 7 Confusion matrix after adversarial defence

| FGSM | | | | | JSMA | | | |
|---|---|---|---|---|---|---|---|---|
| Label | | Predicted | | | Label | | Predicted | |
| | | 0 | 1 | | | | 0 | 1 |
| Actual | 0 | 42110 | 1083 | | Actual | 0 | 41890 | 1303 |
| | 1 | 101 | 47566 | | | 1 | 85 | 47582 |
| PGD | | | | | C&W | | | |
| Label | | Predicted | | | Label | | Predicted | |
| | | 0 | 1 | | | | 0 | 1 |
| Actual | 0 | 42099 | 1094 | | Actual | 0 | 42286 | 907 |
| | 1 | 102 | 47565 | | | 1 | 24933 | 22734 |

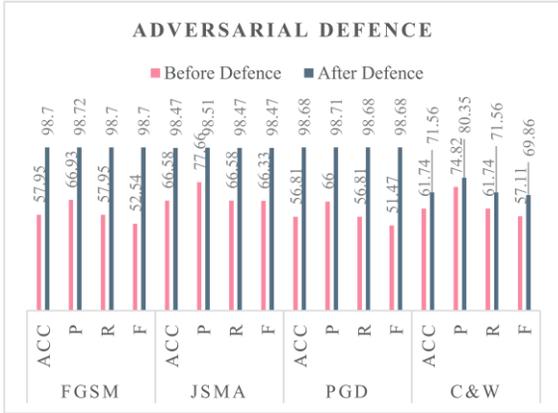

Figure 11. Original vs adversarial defence classification report

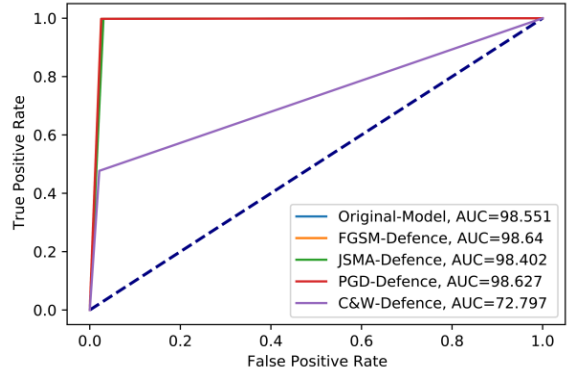

Figure 12. Original vs adversarial defence ROC curve

### 6.2.3.2. Results after gaussian data augmentation defence

In the case of gaussian data augmentation defence, the model is built by adding the noisy counterpart in the dataset. The resulting dataset is either extended data containing the original sample or just the noisy counterparts to the original samples. In our evaluation, the gaussian classifier is built with the sigma and ratio parameters, both are set to 0.01. The clip values of the output vector are -0.98 to +0.99 with the apply predict parameter set to True. The build classifier improved the overall performance of the NIDS model under the four types of adversarial attack situations, i.e., FGSM, JSMA, PGD and C&W. The obtained results of the NIDS model in terms of classification reports, confusion matrix and AUC score are summarised in Table 8, Table 9, and Figure 13. For simplicity and understanding, Figure 14 demographically compares the classification report of the original model in an ideal situation and under the adversarial attack situation in all four cases, i.e., FSGM, JSMA, PGD and C&W.

The improved weighted accuracy, precision, recall and f-score and AUC score, under the FGSM, JSMA, PGD and C&W adversarial attack are (95.71%, 95.78%, 95.71%, 95.71%, 95.793), (97.3%, 97.3%, 97.3%, 97.3%, 97.279), (95.82%, 95.88%, 95.82%, 95.82%, 95.896) and (74.11%, 81.54%, 4.11%, 72.9%, 75.223), respectively. The improved classification report shows that false positive and false negative scenarios, under the four approaches, have been reduced to (1100, 2797), (1332, 1124), (1088, 2712) and (946, 22577), respectively, after applying the gaussian defence.

Table 8 Classification report after gaussian defence

| Label | FGSM AC(%) | P(%) | R(%) | F(%) | JSMA AC(%) | P(%) | R(%) | F(%) |
|---|---|---|---|---|---|---|---|---|
| 0 | | 93.77 | 97.45 | 95.58 | | 97.39 | 96.92 | 97.15 |
| 1 | 95.71 | 97.61 | 94.13 | 95.84 | 97.3 | 97.22 | 97.64 | 97.43 |
| MA | | 95.69 | 95.79 | 95.71 | | 97.3 | 97.28 | 97.29 |
| WA | | 95.78 | 95.71 | 95.71 | | 97.3 | 97.3 | 97.3 |
| Label | PGD | | | | C&W | | | |
| 0 | | 93.95 | 97.48 | 95.68 | | 65.17 | 97.81 | 78.22 |
| 1 | 95.82 | 97.64 | 94.31 | 95.94 | 74.11 | 96.37 | 52.64 | 68.08 |
| MA | | 95.79 | 95.9 | 95.81 | | 80.77 | 75.22 | 73.15 |
| WA | | 95.88 | 95.82 | 95.82 | | 81.54 | 74.11 | 72.9 |

Table 9 Confusion matrix after gaussian defence

| FGSM LB | | Predicted 0 | 1 | JSMA LB | | Predicted 0 | 1 |
|---|---|---|---|---|---|---|---|
| Actual | 0 | 42093 | 1100 | Actual | 0 | 41861 | 1332 |
| | 1 | 2797 | 44870 | | 1 | 1124 | 46543 |
| PGD LB | | Predicted 0 | 1 | C&W LB | | Predicted 0 | 1 |
| Actual | 0 | 42105 | 1088 | Actual | 0 | 42247 | 946 |
| | 1 | 2712 | 44955 | | 1 | 22577 | 25090 |

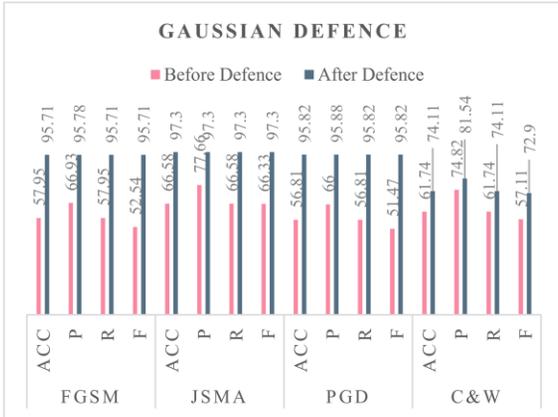

Figure 13. Original vs gaussian defence classification report

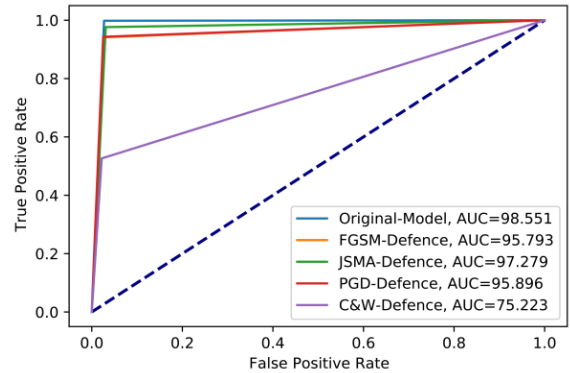

Figure 14. Original vs gaussian defence ROC curve

*6.2.3.3. Results after high confidence defence*

In the high confidence defence [66], the model is rebuilt with the cutoff parameter of 0.05 and apply_predict set to True. The built model is re-evaluated with previously adversarial perturbed examples. The results in terms of classification report, confusion matrix and AUC score are further improved over the four FGSM, JSMA, PGD and C&W adversarial attack algorithms and are summarised in Table 10, Table 11, and Figure 15. Figure 16 demographically compares the classification report of the original model in an ideal situation and under the adversarial attack situation in all four cases, i.e., FSGM, JSMA, PGD and C&W.

The improved weighted accuracy, precision, recall, f-score and AUC score under the FGSM, JSMA, PGD, and C&W adversarial attack are (98.7%, 98.72%, 98.7%, 98.7%, 98.643), (98.47%, 98.51%, 98.47%, 98.47%, 98.402), (98.68%, 98.71%, 98.68%, 98.68%, 98.628) and (71.52%, 80.33%, 71.52%,69.81%, 72.755), respectively. In the case of high Confidence defence, the improved classification report shows that false positive and false negative scenarios, under the four approaches, have been reduced to (1303, 101), (1303, 85), (1093, 102) and (907, 24973), respectively.

Table 10 Classification report after high confidence defence

| Label | FGSM | | | | JSMA | | | |
|---|---|---|---|---|---|---|---|---|
| | AC(%) | P(%) | R(%) | F(%) | AC(%) | P(%) | R(%) | F(%) |
| 0 | 98.7 | 99.76 | 97.5 | 98.62 | 98.47 | 99.8 | 96.98 | 98.37 |
| 1 | | 97.78 | 99.79 | 98.77 | | 97.33 | 99.82 | 98.56 |
| MA | | 98.77 | 98.64 | 98.69 | | 98.57 | 98.4 | 98.47 |
| WA | | 98.72 | 98.7 | 98.7 | | 98.51 | 98.47 | 98.47 |
| Label | PGD | | | | C&W | | | |
| 0 | 98.68 | 99.76 | 97.47 | 98.6 | 71.52 | 62.87 | 97.9 | 76.57 |
| 1 | | 97.75 | 99.79 | 98.76 | | 96.16 | 47.61 | 63.69 |
| MA | | 98.76 | 98.63 | 98.68 | | 79.51 | 72.75 | 70.13 |
| WA | | 98.71 | 98.68 | 98.68 | | 80.33 | 71.52 | 69.81 |

Table 11 Confusion matrix after high confidence defence

| FGSM | | | | JSMA | | | |
|---|---|---|---|---|---|---|---|
| LB | | Predicted | | LB | | Predicted | |
| | | 0 | 1 | | | 0 | 1 |
| Actual | 0 | 42112 | 1303 | Actual | 0 | 41890 | 1303 |
| | 1 | 101 | 47566 | | 1 | 85 | 47582 |
| PGD | | | | C&W | | | |
| LB | | Predicted | | LB | | Predicted | |
| | | 0 | 1 | | | 0 | 1 |
| Actual | 0 | 42100 | 1093 | Actual | 0 | 42286 | 907 |
| | 1 | 102 | 47565 | | 1 | 24973 | 22694 |

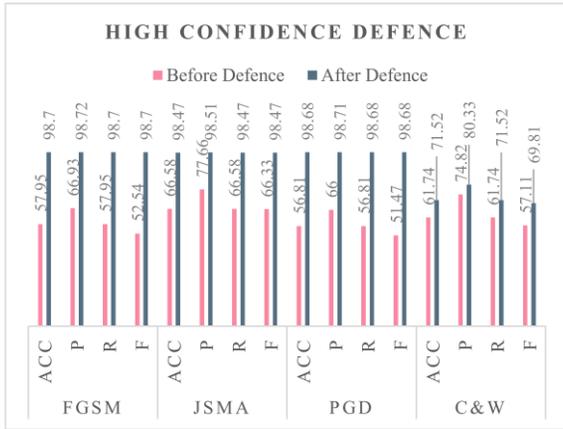
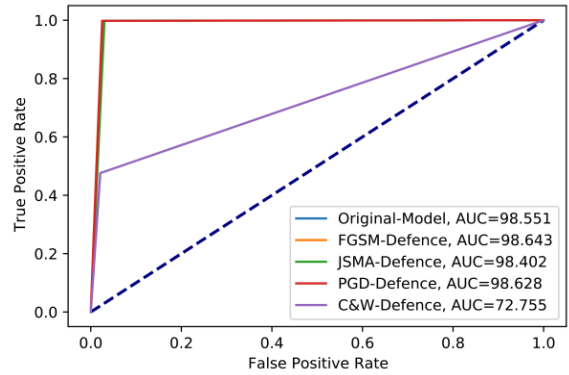

Figure 15. Original vs high confidence defence classification report

Figure 16. Original vs high confidence defence ROC curve

## 6.3. Discussion

The following key findings and conclusions have been drawn from the extensive experimentation and detailed analysis of the four adversarial attacks and three heuristics defence strategies.

C&W attack has proven to be more powerful in most state-of-the-art adversarial attacks. However, in the case of the intrusion detection system, the obtained reduced classification report and confusion matrix are quite similar compared to other powerful attack algorithms, i.e., FGSM, JSMA and PGD. The AUC score under the C&W attack is 63.41, which is significantly high compared to FGSM, PGD (59.232, 58.485) and less with JSMA (68.037).

The effects of adversarial training and high confidence defence strategies are similar for FGSM, JSMA and PGD attacks, as shown in Figure 17. However, in C&W attack, the performance does not improve as much compared to FGSM, JSMA and PGD, in all three adversarial defence strategies. The achieved confidence score in terms of accuracy, precision, recall, f-score, and AUC score for the C&W attack after the adversarial training, gaussian data augmentation and high confidence are (71.56%, 80.35%, 71.56%,69.86%, 72.797), (74.11%, 81.54%, 4.11%, 72.9%, 75.223) and (71.52%, 80.33%, 71.52%,69.81%, 72.755), respectively. It is clearly visible that the maximum reach of the improved performance is up to 80% only in the C&W attack. On the contrary, the same reach is up to 98% in FGSM, JSMA and PGD in all three cases.

The other key point we can conclude is that the defence approach of the Gaussian method outperforms the other two defence methods in the C&W attack. However, in the case of FGSM, JSMA and PGD attack scenarios, the adversarial training and high confidence defence outperform. In future study, we may also extend the same work to see the effects of other adversarial defence approaches over the C&W attack to increase the robustness of the model up to a significant level.

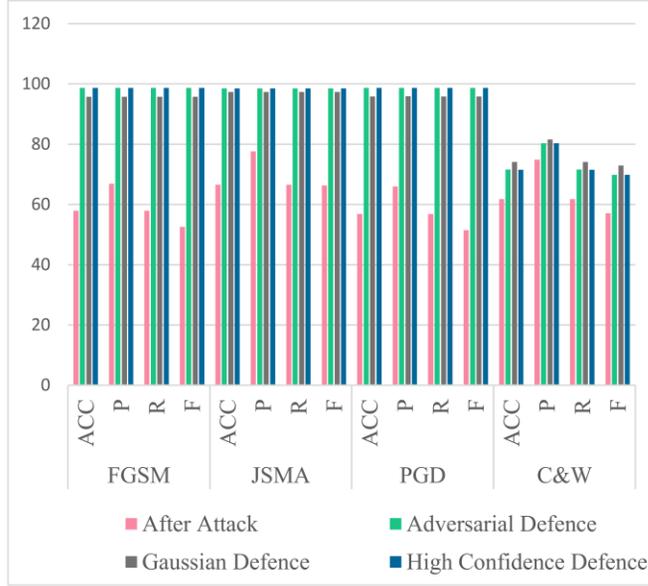

Figure 17. Comparison of three defence methods

The other point of discussion is that we have examined the effects of the adversarial perturbed examples in the constant and known DL-based NIDS model; however, with the transferability concept, it is possible to convert the black-box attack method into the white-box in some cases, as it is possible for the adversary to use a self-made model that mimics the target model, train the model with a similar dataset on which the targeted model is trained. And then generate adversarial examples based on the dummy model to fool the actual targeted model. Therefore, the white-box attack examined here is relevant for a pure black-box attack as well. The research study by Madry et al. [30] revealed that the robustness of the model could be increased by increasing the capacity of the network. The high-capacity model also decreased the problem of saddle points and transferability issues.

**7. Limitation and future scope**

While this study provides valuable insights into the realm of Network Intrusion Detection Systems (NIDS) and adversarial attacks, it is essential to acknowledge its inherent limitations:

- **Adversarial Landscape Complexity:** The study primarily focused on four well-known adversarial attack techniques. However, the adversarial landscape is vast and continuously evolving. Limiting the analysis to these specific attacks may not encompass the full spectrum of threats encountered in practical scenarios.

- **Static Dataset:** The experimentation utilized static datasets, which may not capture the dynamic and evolving nature of network traffic and adversarial behaviors. Real-world NIDS systems operate in dynamic environments, and the study's findings might not fully align with such operational complexities.

- **Simplified Attack Models:** The study assumes knowledge of the NIDS architecture and model details, akin to a white-box scenario. In practice, attackers often operate under more restrictive black-box conditions, which can yield different attack strategies and outcomes.

- **Adversarial Defense Evaluation:** Although the study explored various adversarial defense strategies, it did not conduct an exhaustive analysis of all possible defense mechanisms. Alternative defense approaches may yield different results, and further research is needed in this area.

- **Evaluation Metrics:** The study primarily relied on standard evaluation metrics like accuracy, precision, recall, and F1-score. These metrics, while informative, may not fully encapsulate the impact of adversarial attacks on NIDS systems. Additional metrics and real-world testing could provide a more comprehensive assessment.

**Future Scope of the Study:**

Building upon the insights gained from this study, several avenues for future research and development emerge:

- **Exploration of Emerging Attack Techniques:** As new adversarial attack techniques continue to emerge, future research can investigate their impact on NIDS systems. Staying abreast of evolving attack strategies is crucial for enhancing NIDS robustness.

- **Robust Adversarial-Proof Framework:** There is a compelling scope for the development of a novel and robust adversarial-proof framework tailored explicitly for NIDS. This framework should transcend known attacks and adapt to emerging threats, bolstering NIDS systems against adversarial intrusions [67] [68].

- **Black-Box Attack Analysis:** Future research should conduct in-depth investigations into black-box attacks to overcome the limitations related to simplified model attacks. This involves analyzing scenarios where adversaries possess limited knowledge of the NIDS model. Examining vulnerabilities and susceptibilities in these situations will enable the development of more robust countermeasures [69].

- **Integration of Explainability:** The integration of explainability and model interpretability into NIDS models is a promising avenue. Transparent model predictions empower analysts to identify adversarial incursions more swiftly and formulate effective responses.

- **Concept Drift Adaptation:** Adapting NIDS systems to concept drift in network traffic patterns and attack methodologies is critical. These adaptive systems can detect and respond effectively to shifting threats in dynamic environments.

- **Transfer Learning for Enhanced Robustness:** Leveraging transfer learning to enhance the robustness of NIDS models by transferring knowledge from related domains presents an exciting research direction. Such models can improve threat detection accuracy.

- **Real-World Deployment:** Validation of NIDS defenses through real-world deployment and testing is imperative. Field testing will provide insights into real-time performance and practical utility within authentic network environments.

- **Hybrid Defence Strategies:** Heuristic based defences excel at detecting novel and previously unseen threats, while certified defences are effective against known threats. By combining the two, we can achieve comprehensive threat coverage, reducing the risk of both false negatives (missed threats) and false positives (false alarms). Hence, it would be an excellent research direction to explore.

## 8. Conclusion

This research work demonstrates that no domain, whether constrained or unconstrained, is secure from adversarial attacks. This idea is validated in NIDS by investigating the effects of four adversarial attack techniques, namely, FGSM, JSMA, PGD and C&W. The main aim is to tweak DL based NIDS to misclassify the network anomalies into benign network traffic by injecting adversarial perturbed examples into the system. As a result, it is seen that the performance of the NIDS has been significantly degraded in terms of accuracy, precision, recall, f-score and AUC. To mitigate this effect and increase the robustness of the model, we further investigated three adversarial defence strategies, i.e., Adversarial Training, Gaussian Data Augmentation and High Confidence. The confidence score of the model has been significantly improved in all four adversarial attack scenarios after the defence implementation.

To the best of our knowledge, this research article provides an in-depth analysis of various state-of-the-art adversarial attacks and their defence mechanism in the constraints domain (NIDS) with extensive experimentation. The obtained results recommend avoiding DL-based NIDS for mission-critical and real-time streaming applications or deploying it with a proper and strong defence mechanism to prevent any adversarial attacks. The other key point is that adversaries have the potential to modify most of the features in the network traffic without altering the functioning of the network behaviour, making intrusion more challenging to detect. In future, we aim to investigate the effect of the proposed approach on other ML and DL architectures. The same approach can also be extended to investigate the transferability concept in the adversarial machine learning domain. It is also recommended to apply the same method over concept drift network streaming data-based NIDS.

## Statements and Declarations

**Funding:** Not Applicable

**Conflicts of Interest:** The authors declare that they have no known competing financial interests or personal relationships that could have appeared to influence the work reported in this paper.

**Authors' contributions :**

Author-1: Conducted experiments and wrote the complete manuscript.

Author-2: Reviewed the manuscript and provided valuable suggestions for further improvements.

Author-3: Suggested the revision process of the manuscript in response to reviewer comments.

**Data and Code Availability Statement**: The datasets analyzed during the current study are available online:

https://www.unb.ca/cic/datasets/ids-2017.html

The code generated and analyzed during the current study will be available on reasonable request.

**Abbreviations:**

All the abbreviations used in this manuscript are listed below.

AC– Accuracy

AML – Adversarial Machine Learning

ANN- Artificial Neural Network

AT – Adversarial Training

AUC ROC- Area Under the ROC Curve

BIM – Basic Iterative Method

C&W- Carlini and Wagner

CICIDS-2017 - Canadian Institute for Cybersecurity Intrusion Detection System 2017

CM – Confusion Matrix

CNN- Convolutional Neural Networks

CR – Classification Report

DL – Deep Learning
DT – Decision Tree

F- F-Score

FGSM- Fast Gradient Sign Method

GAN – Generative Adversarial Attack

GDA – Gaussian Data Augmentation

HC- High Confidence
HSG - Histogram Sketch Generation

IDS - Intrusion Detection System

JSMA- Jacobian Based Saliency Attack

KNN- k-Nearest Neighbor

LB - Label

MA – Macro Average

ML – Machine Learning

NIDS – Network Intrusion Detection System

NN- Neural Network

P – Precision

PGD- Projected Gradient Descent

R -Recall
RF - Random Forest

ROC Curve - Receiver Operating Characteristic Curve

SVM – Support Vector Machine

WA – Weighted Average